\pdfoutput=1

\documentclass[11pt]{article}

\usepackage[preprint]{acl}

\usepackage{times}
\usepackage{latexsym}

\DeclareMathAlphabet\mathbfcal{OMS}{cmsy}{b}{n}
\usepackage{amssymb}
\usepackage{amsmath}
\usepackage{graphicx}
\usepackage{multirow,booktabs}
\usepackage{xcolor}
\usepackage{CJKutf8}
\usepackage{amsthm}
\usepackage{mathrsfs}
\usepackage{hyperref}
\graphicspath{{Images/}}

\usepackage[T1]{fontenc}

\usepackage[utf8]{inputenc}

\usepackage{microtype}

\usepackage{inconsolata}

\title{DPPA: Pruning Method for Large Language Model to Model Merging}

\author{Yaochen Zhu$^1$, Rui Xia$^1$, Jiajun Zhang$^{2}$  \\
        $^1$Nanjing University of Science and Technology, China \\
        $^2$School of Artificial Intelligence, University of Chinese Academy of Sciences\\
        \texttt{\{yczhu, rxia\}@njust.edu.cn, jjzhang@nlpr.ia.ac.cn } }
        
\begin{document}
\maketitle
\begin{abstract}

Model merging is to combine fine-tuned models derived from multiple domains, with the intent of enhancing the model's proficiency across various domains. The principal concern is the resolution of parameter conflicts.
A substantial amount of existing research remedy this issue during the merging stage, with the latest study focusing on resolving this issue throughout the pruning stage. 
The DARE approach has exhibited promising outcomes when applied to a simplistic fine-tuned model. However, the efficacy of this method tends to wane when employed on complex fine-tuned models that show a significant parameter bias relative to the baseline model.
In this paper, we introduce a dual-stage method termed Dynamic Pruning Partition Amplification (DPPA), devised to tackle the challenge of merging complex fine-tuned models. 
Initially, we introduce Dynamically Pruning (DP), an improved approach based on magnitude pruning, which aim is to enhance performance at higher pruning rates.
Subsequently, we propose Dynamically Partition Amplification (DPA), a rescaling strategy, is designed to dynamically amplify parameter partitions in relation to their significance levels.
The experimental results show that our method maintains a mere 20\% of domain-specific parameters and yet delivers a performance comparable to other methodologies that preserve up to 90\% of parameters.
Furthermore, our method displays outstanding performance post-pruning, leading to a significant improvement of nearly 20\% performance in model merging.
We make our code on~\href{https://github.com/northsky0307/DPPA-Pruning-Method-for-Large-Language-Model-to-Model-Merging.git}{Github}.

\end{abstract}
\begin{figure*}
    \centering
    \includegraphics[width=1.0\textwidth]{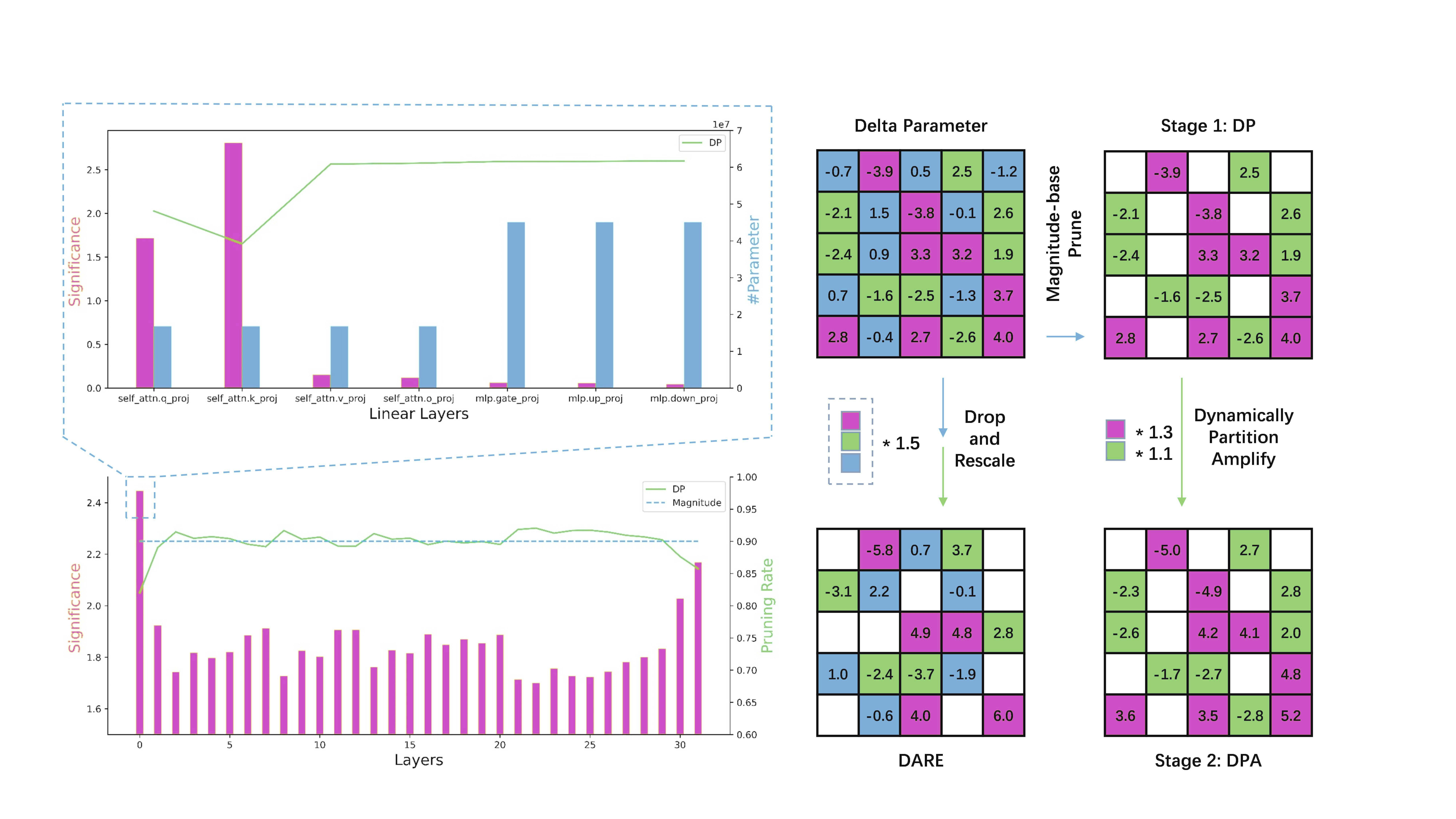}
    \caption{
    Within the diagram's left segment, it is visible that our Dynamic Pruning (DP) technique adaptively modifies the pruning rate at both layer and linear layer levels, distinguishing it from Magnitude Pruning. On the diagram's right segment, we can see the integration of DP and Dynamic Pruning Algorithm (DPA), paralleled with the drop and rescale operations inherent in the DARE system. This integration enhances complex model performance after the pruning process significantly.
    }
    \label{fig:main_method}
\end{figure*}

\section{Introduction}

Model merging, alternatively referred to as model fusion, signifies a method that amalgamates finely-tuned models originating from diverse domains, with the primary objective of enhancing the model's proficiency across numerous domains. The principal issue leading to performance degradation post-merging is predominantly attributable to parameter conflicts. Therefore, the prime focus is the resolution of such parameter conflicts.
To more accurately depict the characteristics of the respective domains, the merging parameter is defined by the discrepancy between the parameters of the SFT model and the base model, known as delta parameters.
Current pruning techniques~\cite{SparseGPT,Wanda}, primarily emphasize on minimizing the parameter count. However, these methods may not produce satisfactory outcomes when implemented on delta parameters.

The predominant methods address the issue of parameter conflicts in the merging stage as highlighted in the works by~\cite{AdaMerging,times,regmean}. However, contemporary research has shifted towards mitigating this problem during the pruning stage.
A reduction in the number of parameters corresponds to a decrease in the incidence of parameter conflicts.
A recent method titled DARE~\cite{mario} presents a practice that involves arbitrary removal and resizing. Such a strategy displays favourable outcomes on basic fine-tuned models. However, its efficacy tends to falter when implemented on models with more significant divergences from the fundamental model parameters. The publication acknowledges this drawback, stating, ``However, once the models are continuously pretrained, the value ranges can grow to around 0.03, making DARE impractical.''
In addition, we hypothesize that models with distinctly noticeable disparities from the basal model parameters typically exhibit superior results post-engaging with a sophisticated fine-tuning procedure.

In this study, we propose a two-stage method called DPPA to address the challenge of fusing complex fine-tuned models.
First, we introduce Dynamically Pruning (DP), an improved approach based on magnitude pruning with the primary intent of boosting performance at higher pruning rates.
Subsequently, we propose Dynamically Partition Amplification (DPA), a rescaling technique that aims to dynamically amplify partitions of parameters based on their varying levels of significance.

Dynamically Pruning (DP) is employed to dynamically adjust the pruning rate based on the importance of different linear layers.
OWL~\cite{OWL} observed that the importance of parameters varies across different layers. We posit that, in circumstances of elevated pruning rates, it is imperative to intensify the refinement of parameter significance and to modify the pruning rate at the linear layer level. For example, As depicted in Figure~\ref{fig:main_method}, the parameters present in layer $0$ of the Delta parameter exhibit higher prominence when compared to those found in layer $22$. Moreover, it is apparent that the Q and K parameters in layer $0$ hold more significant value when compared to other linear layers.
Our methodology involves dissecting the model into different layers, wherein the linear layers (such as Q, K, V, O in Attention and upsampling/downsampling in MLP) are deemed as the lowest units for modulating the pruning rates.

Moreover, Dynamically Partition Amplification (DPA) is a rescale method, which is built upon the pruning approach.
We surmise that parameters displaying significant deviations from the baseline model during fine-tuning are of utmost importance. We allocate priority to this critical subset of parameters and enhance their values. Once the optimal enhancement rate for this subset is determined, we proceed to assess the subsequent subset of parameters in order of importance, and so on.

The base model we employ in our research is LLaMA 2~\cite{llama2}. We carry out fine-tuning across three distinct domains, namely mathematics, finance, and law. The experimental results suggest that our methodology maintains only 20\% of the domain-specific parameters, yet the performance is comparable to alternative methods that hold onto 90\% of said parameters. In addition, our approach, due to its impressive efficacy post-pruning, also demonstrates noticeable improvement, around 20\%, concerning model merging. We further substantiate the viability of DPA on DARE, although it doesn't yield a level of performance equal to DPPA. Nonetheless, it does improves performance moderately. We execute trials in contexts of three-domain and two-domain merging, and the findings suggest that the influence of the additional domain on our method is essentially insignificant.

\section{Related Work}
\subsection{Pruning Technique}
Traditional pruning techniques are a type of model compression that aim to decrease the number of parameters in a model ~\cite{Survey_Model_Compression}. There have been several studies conducted on this topic, both in the era of pretrained language models and before~\cite{NM,RP1,RP2,RP3}. However, progress in these studies has been relatively slow in the era of large language models, as pruning requires a substantial amount of data for fine-tuning, which is not feasible for such models. To tackle this issue, LORA fine-tuning was proposed by \citet{prun_lora} to restore the original performance. Recently, some studies have shifted their focus to pruning methods that do not necessitate fine-tuning. For instance, SparseGPT~\cite{SparseGPT} utilizes the Hessian matrix for pruning and reduces reconstruction error through subsequent weight updates. Wanda~\cite{Wanda} combines weight magnitudes with input activations to retain parameters that better align with the current data distribution. DSOT~\cite{dsot} proposes a parameter adjustment method to minimize the discrepancy between the source model parameters and the pruned model parameters. OWL~\cite{OWL} introduces non-uniform layered sparsity, which is advantageous for higher pruning rates.

\subsection{Special Domain Fine-tune Model}
Since the advent of the machine learning era, models have required adjustments on specific data to achieve desired performance. In the era of pretrained language models, this approach has been slightly modified. Researchers first pretrain a general model and then fine-tune it on domain-specific data, with the primary goal of leveraging the capabilities of the pretrained model. This is even more crucial in the era of large language models, resulting in the development of numerous models in different domains. For example, in the code domain \cite{codellama,codewave,wizardcode}, mathematics domain \cite{wizardmath,mmothmath,ovmmath,toramath,RFTmath}, medical domain \cite{medicalClinical,meditron,meidcalcamel}, and finance domain \cite{financefingpt,financefingpt2,financepixiu}.

Although we have obtained many fine-tuned models in specific domains, if we want a single model to have the capability to handle multiple domains, the fundamental approach is to fine-tune the model on all domain data together. However, this requires a significant amount of computational resources. Therefore, model fusion methods have gained attention.

\subsection{Model Merge}
\label{sec:related_merge}
The mainstream model fusion methods can be divided into four sub-domains: alignment \cite{algin}, model ensemble \cite{model_ensemble}, module connection \cite{model_connect}, and weight averaging \cite{weight_average}. Among these methods, only weight averaging reduces the number of model parameters, while the others require the coexistence of model parameters from multiple domains \cite{modelfusionservey}. Within the weight averaging sub-domain, there are also several approaches, such as subspace weight averaging \cite{subspace}, SWA\cite{SWA}, and task arithmetic \cite{task_arithmetic}. We are particularly interested in the task arithmetic sub-domain because it does not require the fusion of multiple models during the training process. Instead, it only requires obtaining the weights of a fully trained model.

The task arithmetic approach suggests that there is a domain-specific offset between the fine-tuned model weights and the base model weights. By adding or subtracting these offsets from multiple domains, it is possible to fuse or selectively exclude the capabilities of certain domains. Subsequent works have explored the application of task arithmetic to LORA \cite{LORA1,LORA2,LORA3}, as well as how to better fuse models and reduce conflicts between parameters. \citet{rasie_task} achieved this by scaling the coefficients of different models during the fusion process to mitigate conflicts between models. \citet{AdaMerging} further proposed adjusting the scaling coefficients at the model hierarchy level to address conflicts caused during model fusion at a finer granularity. \citet{times} selected which model weights to retain at specific positions by comparing the absolute values of conflicting weights. \citet{regmean} adjusted the entire conflicting vector in vector space to ensure that the L2 distance between this vector and multiple original vectors remains equal.

\subsection{Federated Learning}
Federated learning is a setup where multiple clients collaborate to solve machine learning problems, coordinated by a central aggregator. This setup also allows for decentralized training data to ensure privacy of data on each device \cite{federated}. Model fusion methods naturally possess the ability to combine locally trained models together. Furthermore, since the central aggregator receives locally trained weights, there is no need to worry about data leakage issues.

\section{Methodology}
The purpose of our approach is to integrate multiple fine-tuned models from various domains into a single model. Therefore, we first review the definition of model merging.

Our approach consists of four parts, as shown in Fig.~\ref{fig:main_method}. 
First, we calculate the delta parameter, signifying the weight disparity between the fine-tuned models and the Base model. Second, we implement a variant of the magnitude pruning technique, referred to as DP, which personifies superior performance at elevated pruning rates. This technique prunes the delta parameter to mitigate conflicts in the parameter space during model integration. Subsequently, we introduce a rescaling method, DPA, to amplify the pruned delta parameter, resulting in enhanced performance. Conclusively, we amalgamate the parameter from various fine-tuned models and incorporate them into the Base model, thus yielding a single model with multidomain capabilities.

\subsection{Model Merging Problem}
The purpose of model Merging is to enhance the capability of a single model by combining fine-tuned models from multiple domains. Specifically, for fine-tuned models $M^1 \sim M^k$, each associated with different domains $D^1 \sim D^k$, where each domain comprises a set of tasks $D^i = \{T^i_1 \sim T^i_n\}$. 
Here, $k$ represent the number of domain, $i$ represents a specific domain, and $n$ represents the number of tasks within that domain.

By merging $M^1 \sim M^k$, we obtain the integrated model $M^m$, which possesses the ability to handle tasks from $D^1 \sim D^k$ simultaneously.

\subsection{Delta Parameter}
For each fine-tuned model in each domain, we can find the corresponding pre-trained model, known as the Base model $M^B$. For domain $i$, we have the weights $W^i$ of the fine-tuned model $M^i$ and the weights $W^B$ of the base model. We define the delta parameter as the transition of the parameter space distribution from the base model to the fine-tuned model, represented as $\Delta^i = W^B - W^i$. Conducting an analysis on the delta parameter facilitates a more comprehensive comprehension of the alterations introduced by the fine-tuning process.

\subsection{DPPA}
First, we introduce Dynamically Pruning (DP), an improved approach based on magnitude pruning which aim is to enhance performance at higher pruning rates.
Subsequently, we propose Dynamically Partition Amplification (DPA), a rescaling technique that aims to dynamically amplify partitions of parameters based on their varying levels of significance.

\subsubsection{DP: Dynamically Pruning}
We propose to use linear layers as the minimum unit and adjust the pruning rate based on the significance of different linear layers. Here, the linear layers, such as Q, K, V, O in Attention and up/down sampling in MLP, are more fine-grained units compared to model layers. We first describe how to define the significance of parameters and then explain the method for adjusting the pruning rate.

Within the framework of OWL~\cite{OWL}, significance of a parameter is defined as the value exceeding the average weight magnitude by N-fold. Drawing inspiration from OWL, we have redefined importance.
Rather than depending on the quantity of parameters, it now considers the accumulated magnitudes of parameters that surpass the average magnitude by a factor of N. 
This refinement includes more comprehensive information about weight parameters. Based on empirical findings from past studies, we set N to 5. This approach allows us to determine the significance of parameters on both the model layer and the linear layer levels.

Once the significance of the parameters has been determined, we can adjust the pruning rate accordingly. Following the principle that higher parameter importance corresponds to lower pruning rates, we define the pruning rate fluctuation at the model level as:
\begin{equation}
dif(\Delta_l) = - sig(\Delta_l) + \frac{1}{n}\sum_{l=1}^{n} {sig(\Delta_l)}
\end{equation}
where $dif$ represents the difference between importance and its mean, for briefly, we reduce domain-specific $\Delta^i$ to $\Delta$, thus $\Delta_l$ represents paremeters in model layer $l$, $sig()$ represents significance of the parameter, $n$ represents the number of model layers, respectively.

Furthermore, since the number of parameters in different linear layers may vary, we introduce a weighting factor for the parameter importance, as shown:

\begin{equation}
mean(\Delta_{lj}) =  \frac
{\sum_{l=1}^{n} \sum_{j=1}^{m} sig(\Delta_{lj})* \left\|\Delta_{lj}\right\|_0}
{\sum_{l=1}^{n} \sum_{j=1}^{m} \left\|\Delta_{lj}\right\|_0}
\end{equation}
\begin{equation}
dif'(\Delta_{lj}) = - sig(\Delta_{lj}) + mean(\Delta_{lj}),
\end{equation}
where $\Delta_{lj}$ represents paremeters in model layer $l$ linear layer $j$, $m$ represents the number of linear layers in model layer, $\left\|X\right\|_0$ represents the parameter count of $X$, respectively.

Finally, we define the maximum value of pruning rate fluctuation, denoted as $\lambda$, based on previous experimental findings, and set it to 0.08. By considering both the fluctuation within linearlayer-level and layer-level, we derive the final pruning rate for each linear layer as follows:
\begin{equation}
norm(x) = \frac{x * \lambda}{max \, abs(x)}
\end{equation}
\begin{equation}
{\Theta_{lj}} = \alpha + norm(dif(\Delta_l)) + norm(dif'(\Delta_{lj})),
\end{equation}
where $\alpha$ represents original pruning rates, $abs$ represents absolute value.

\subsubsection{DPA: Dynamically Partition Amplification}
\begin{figure}
    \centering
    \includegraphics[width=0.5\textwidth]{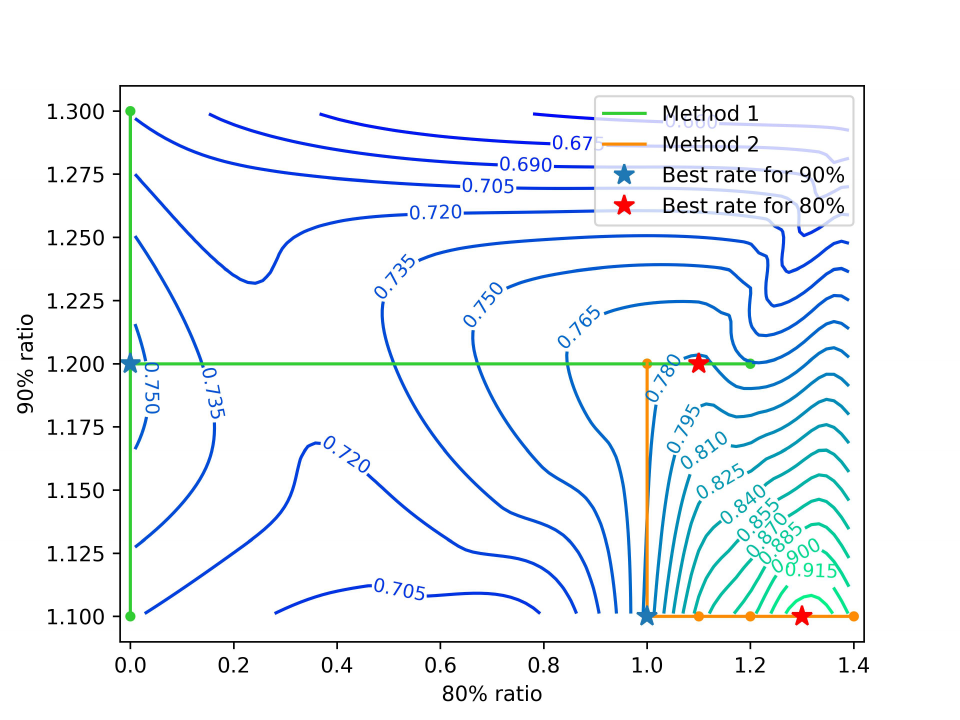}
    \caption{
    We utilize \textcolor[RGB]{50,205,51}{green} and \textcolor[RGB]{255,139,0}{orange} lines to represent the trajectories of amplification rate search. Among them, the \textcolor[RGB]{31,120,180}{blue star} represents the optimal rate searched at a 90\% pruning parameter, while the \textcolor[RGB]{254,0,0}{red star} represents the optimal rate searched at an 80\% pruning parameter. The contour lines depict the specific performance in the mathematical domain.}
    \label{fig:search_path}
\end{figure}

After DP, we obtain the pruned delta parameters at various pruning rates. Our goal moving forward is to enhance performance while ensuring a consistent pruning rate.
The model's performance clearly demonstrates an initial rise followed by a gradual decline as the scaling rate augments. This pattern is consistently observed across various pruning rates, as illustrated in Fig.~\ref{fig:search_path}. Moreover, we postulate that during the fine-tuning stage, parameters with substantial deviations significantly influence their corresponding fields.

Therefore, we propose DPA, a methodology that segments parameters at varying pruning rates and dynamically alters the enhancement factors for each division. We consider two methods of initialization to accomplish this dynamic adaptation, with the ultimate goal of locating the optimum results. Ultimately, our selection of the most effective method will be determined by the results.

\paragraph{Method 1} We adjust the parameters in the 90\% pruning rate partition by setting the rest to zero. 
We surmise that partitions with elevated pruning rates hold a greater degree of importance. Consequently, the precedence in sorting partitions is primarily influenced by their respective pruning rates. Illustratively, the parameters within the 90\% pruning rate section are perceived as having a higher value compared to those within the 80\% pruning rate partition. Upon the acquisition of the ideal amplification ratio, we progressively incorporate parameters from the 80\% pruning rate partition, scaling only the newly included parameters. This method persists until the attainment of the target pruning rate.
The resulting curve of this method is illustrated by the \textcolor[RGB]{50,205,51}{green} line in Fig.~\ref{fig:search_path}.

\paragraph{Method 2}
We employ the partition that aligns with the target pruning rate directly during the adjustment of the 90\% partition.
We recognize that Method 1 could generate excessively large amplification factors for more significant partitions, thereby causing a substantial displacement in the parameter space of partitions with lower pruning rates. This shift may ultimately decrease performance when integrating parameters from partitions with lower pruning rates.
In this strategy, when modifying more critical partitions, we consider the parameter distribution of less significant partitions. Method 1 is outperformed by this method when the pruning rate aim is low.
The resulting curve of this method is illustrated by the \textcolor[RGB]{255,139,0}{orange} line in Fig.~\ref{fig:search_path}.

\subsection{Model Merging with DPPA}
After applying DPPA, we are merely required to integrate parameters derived from distinct models. In Section~\ref{sec:related_merge}, we referred to multiple existing methodologies for model fusion. However, our primary objective is to enhance the pruning technique. As such, we employ AdaMerging~\cite{AdaMerging}, a state-of-the-art merging approach, to confirm the parameter integration following the pruning process. It warrants mentioning that models destined for merging via fine-tuning ought to originate from an identical pre-trained model, as existing fusion techniques do not support the integration of heterogeneous models.

Thus, we get the final merging model:
\begin{equation}
{W^m} = W^B + \Sigma_{i=1}^k \text{DPPA}(\Delta^i)
\end{equation}

\section{Experiments}

\subsection{Experimental Setup}

\paragraph{Pre-Trained Backbone and Fine-tune Models}
We have taken into consideration the need to fine-tune the same base model for different domains and the impact of the base model's performance. Therefore, we have decided to choose LLaMa 2\cite{llama2} as the base model, instead of LLaMa\cite{llama}, Mistral\cite{mistral}, or other pre-trained models. For the three domains, mathematics, finance and law, we have selected three models with good performance, namely Abel\cite{abel}, Finance-chat and Law-chat\cite{adaptllm}.

\vspace{-0.2cm}
\paragraph{Datasets}
\label{sec:Domain_Ratio}
For each domain, we have chosen two datasets. In the mathematics domain, we have selected GSM8k\cite{gsm8k} and MATH\cite{math}. We evaluate the models' performance using zero-shot accuracy and utilize the testing script provided by Abel\cite{abel}. As for the finance domain, we have chosen FiQA\_SA\cite{FISA_QA} and FPB\cite{FPB}. As for the law domain, we have chosen SCOTUS~\cite{supreme-court-database} and the UNFAIR\_ToS~\cite{unfair}. Similarly, we evaluate the models' performance using zero-shot accuracy. 
Since AdaptLLM\cite{adaptllm} does not provide a testing script, we consider the multiple-choice question to be correct when the predicted sentence contains the correct choice.

\paragraph{Evaluation Metric}
To evaluate the correlation between the pruned and fine-tuned pruned model, we formulated the Task-Ratio metric. Furthermore, to exhibit the model's generalization proficiency within each domain, we opted for two datasets. We established the Domain-Ratio as a measure for gauging the specialized capability of the pruned model within a particular domain. The formula for Domain-Accuracy is as follows:
\begin{equation}
\text{Task-Ratio}_j = \frac{R(M_{pruned}, T_j)}{R(M_{dense}, T_j)}
\end{equation}
\begin{equation}
\text{Domain-Ratio} =\sqrt[n]{ \Pi^{n}_{j=1} \text{Task-Ratio}_j},
\end{equation}
where $R(M,T)$ represents the performance of model $M$ on task $T$, $M_{dense}$ refers to the fine-tuned model, $M_{pruned}$ represents the pruned model, and $T_j$ represents task $j$ within the given domain, respectively.

\paragraph{Implementation Details}
In our study, we employed the vLLM framework for reasoning. For the datasets GSM8k and MATH, we set the batch size to 32. As for the FiQA\_SA, FPB, SCOTUS and UNFAIR\_ToS datasets, we set the batch size to 1. We utilized a greedy decoding approach with a temperature of 0. The maximum generation length for all tasks was set to 2048. Our experiments were conducted using the NVIDIA Tesla A100 GPU.

\subsection{Baseline Method}
We establish two methods of pruning-base, and one of randomly deleting and scaling as baseline. they are described below:
\begin{itemize}
  \item \textbf{Magnitude} \cite{magnitude} sorts weights based on their absolute values, keeping weights with larger absolute values and removing weights with smaller absolute values.
  \item \textbf{OWL} \cite{OWL} building upon magnitude pruning, this method considers that parameter importance varies across different layers of the model.
  \item \textbf{DARE} \cite{mario} suggests that after pruning, the sum of parameter values should remain the same. Therefore, it initially performs random pruning and then expands the remaining parameters based on the pruning rate to achieve the original sum of parameter values.
\end{itemize}

\subsection{Main Result of DPPA}
\begin{table}
\centering
\small 
\resizebox{0.5\textwidth}{!}{
\begin{tabular}{lccccc}
\toprule
{\textbf{Sparse ratio}} &{Magnitude} &{OWL}  & {DARE} & {DPPA} \\
\midrule
{Math-Dense}   \\
{10\%} & {96.46} & {96.69} & {96.64} & {-}  \\
{80\%} & {80.12} & {77.11} & {87.41} & {\textbf{97.08}} \\
{90\%} & {53.41} & {54.09} & {73.44} & {\textbf{86.85}} \\
\midrule
{Fin-Dense}  \\
{10\%} & {90.81} & {89.12} & {91.04} & {-} \\
{80\%} & {71.04} & {74.92} & {84.01} & {\textbf{96.65}}\\
{90\%} & {54.71} & {56.74} & {82.90} & {\textbf{92.11}}\\
\midrule
{Law-Dense}  \\
{10\%} & {95.74} & {110.74} & {116.02} & {-} \\
{80\%} & {113.98} & {\textbf{124.97}} & {79.93} & {116.02}\\
{90\%} & {84.35} & {\textbf{121.42}} & {69.33} & {110.55}\\
\bottomrule
\end{tabular}
}
\caption{
Domain-Ratio of different pruning methods at various pruning rates. Additional results under different pruning rates and the performance on a single dataset are presented in Appendix~\ref{sec:specific_tasks}.
}
\label{tab:main_result_of_pruning}
\end{table}
The results of the pruning methods are shown in Table~\ref{tab:main_result_of_pruning}. We compare the results of DPPA with two magnitude-based pruning methods, as well as compare the results of DARE.
The experimental results show that our approach retains only 20\% of the specific domain parameters, yet achieves comparable performance to other methods that retain 90\% of the specific domain parameters.
Due to space limitation, we place the completed experimental table in Appendix~\ref{sec:specific_tasks}.

\subsection{Abnormal Situations in Law Domain}
We believe that our method can achieve performance levels as close as possible to the dense model itself. However, for some tasks that require performance beyond what the dense model can offer, our method may not be as effective. 
In contrast to the expected results from normal pruning, in the law domain, the pruned models significantly outperformed the dense model. The best performance was observed in the range of 120-140\% of the dense model's performance, as pruning rates varied from 10\% to 90\%. We attribute this phenomenon to two factors: first, the relatively low performance of the law domain finetune model itself, and second, the possibility that the model was in a local minimum, causing any offset introduced by pruning to enhance the model's performance.

\begin{table}
\centering
\small 
\resizebox{0.38\textwidth}{!}{
\begin{tabular}{lccc}
\toprule
{\textbf{Domains}} &{Magnitude} &{OWL} &{DP}\\
\midrule
{Math} &{53.41} & {54.09} & {\textbf{54.97}} \\
{Fin} & {54.71} & {56.74} & {\textbf{62.06}} \\
{Law} & {84.35} & {\textbf{121.42}} & {110.55}\\
\bottomrule
\end{tabular}
}
\caption{
Domain-Ratio of DP at a pruning rate of 90\%.
}
\label{tab:dp}
\end{table}
\subsection{The Effectiveness of DP}
As shown in Table~\ref{tab:dp}, DP to achieve better performance at high pruning rates.
This is because DP adjusts the significance of linear layer parameters within each layer, allowing for the retention of more crucial parameters at high pruning rates.

\subsection{The Generality of DPA}

\begin{table}
\centering
\small 
\resizebox{0.42\textwidth}{!}{
\begin{tabular}{lccc}
\toprule
{\textbf{Domains}} &{DARE} &{DARE+DPA} & {DPPA} \\
\midrule
{Math} & {73.44} & {83.63} & {\textbf{86.85}} \\
{Fin} & {82.90} & {85.08} & {\textbf{92.11}} \\
{Law} & {69.33} & {\textbf{120.89}} & {110.55} \\
\bottomrule
\end{tabular}
}
\caption{
Domain-Ratio of DARE using DPA at a pruning rate of 90\%.
}
\label{tab:mario_with_drop}
\end{table}

We investigated the generality of the DPA method by applying it to the state-of-the-art method, DARE. Considering that the DARE method already amplifies the parameters and achieves significant amplification at high pruning rates (5 times for 80\% and 10 times for 90\%), we modified the approach to dynamic reduction instead. Following the methodology, we conducted experiments, and the results are presented in Table~\ref{tab:mario_with_drop}. 

\begin{figure*}
    \centering
    \includegraphics[width=1\textwidth]{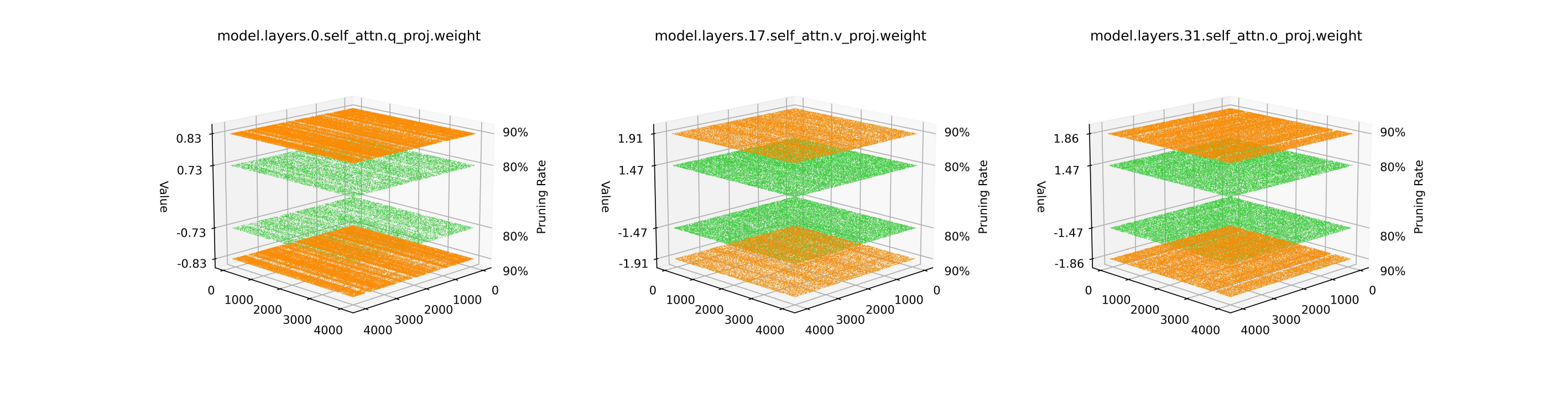}
    \caption{
    After analyzing the pruned parameters of the financial model, it is evident that there is a higher parameter count in the initial and final 0, 31 layers, while the middle 17 layers have fewer parameters. Additionally, in the Q, K, V components, it is observed that 90\% of the parameters are concentrated in certain dimensions. To facilitate observation, we have amplified the value by a factor of 1000.}
    \label{fig:parameter_analysis}
\end{figure*}

\subsubsection{When can DP replace DARE?}

\begin{table}
\centering
\small 
\resizebox{0.5\textwidth}{!}{
\begin{tabular}{lccccc}
\toprule
{\textbf{Model}} &{Min} &{10\%}& {90\%} & {Max} \\
\midrule
{Math-Dense} & {-0.01733} & { -0.00114} & {0.00114} & {0.02014} \\
{Fin-Dense} & {-0.02612} & {-0.00160} & {0.00160} & {0.02011} \\
{Law-Dense} & {-0.02185} & {-0.00158} & {0.00158} & {0.02027} \\
\bottomrule
\end{tabular}
}
\caption{
The offset of different models from the base model at different position proportions.
}
\label{tab:amplitude}
\end{table}
According to the DARE paper, the method's performance is not satisfactory when the parameter deviation from the base model exceeds 0.03. Our observations indicate that the larger the offset, the poorer the performance. This is evident from the model offset presented in Table~\ref{tab:amplitude}.
Certainly, we will present more comprehensive results in Appendix~\ref{sec:model_offset}.
When DARE falls below 90\% performance at a pruning rate of 90\%, our method can serve as a viable alternative.

\subsection{Why DPPA is Useful?}
To investigate this question, we conducted an analysis of the Delta parameters, as shown in Fig~\ref{fig:parameter_analysis}. We explored the relationship between the remaining parameters after DP at different pruning rates and different linear layers. The graph indicates that although DP is an unstructured pruning method, it exhibits some characteristics of structured pruning in the results of high pruning rates for the Delta parameters.
This dimension partitioning provides some interpretability for the distribution of parameter space in specific domains.
Therefore, when we use DPA, by amplifying the parameters, we strengthen the weights of the domain in these dimensions and restore certain capabilities.

\subsection{Main Result of Merge Methods}
\begin{table}
\centering
\small 
\resizebox{0.5\textwidth}{!}{
\begin{tabular}{lccc}
\toprule
{\textbf{Method \& Pruning Rate}} &{Math}&{Fin}&{Law}\\
\midrule
{DARE 90\%}& {7.89} & {51.48}& {53.86} \\
{DPPA 90\%}& {89.95} & {85.24}& {122.08} \\
\midrule
{DARE 80\%}& {32.61} & {74.49}& {78.11} \\
{DPPA 80\%}& {\textbf{91.28}} & {\textbf{95.20}}& {\textbf{146.23}} \\
\bottomrule
\end{tabular}
}
\caption{
Domain-Ratio of the model that combines domains mathematics, finance and law.}
\label{tab:model_with_three_merge}
\end{table}
\begin{table}
\centering
\small 
\resizebox{0.40\textwidth}{!}{
\begin{tabular}{lcc}
\toprule
{\textbf{Method \& Pruning Rate}} &{Math}&{Fin}\\
\midrule
{DARE 90\%}& {21.10} & {64.88} \\
{DPPA 90\%}& {89.25}& {79.40} \\
\midrule
{DARE 80\%}& {58.43} & {77.16} \\
{DPPA 80\%}& {\textbf{92.75}}& {\textbf{95.45}} \\
\bottomrule
\end{tabular}
}
\caption{
Domain-Ratio of the model that combines domains mathematics and finance.}
\label{tab:main_result_with_merge}
\end{table}
We validate the effectiveness of our pruning method for the task of model fusion by integrating models.
In Table~\ref{tab:model_with_three_merge}, we present the merging results for three domains, while in Table~\ref{tab:main_result_with_merge}, we showcase the merging results for two domains.
We choose pruning rates of 80\% and 90\% to compare the results of model merging, as shown in the Table~\ref{tab:main_result_with_merge}.
Based on the results, our method demonstrates an improvement of nearly 20\% in performance compared to DARE at the same pruning rate. This finding substantiates the efficacy of our pruning approach in the context of complex model fusion.

By comparing the results in Table ~\ref{tab:model_with_three_merge} and Table ~\ref{tab:main_result_with_merge}, It can be observed that the integration of a fine-tuned model from an additional domain considerably influences DARE's performance, causing significant performance deterioration.
In comparison, our method achieves comparable performance.
Upon augmenting an additional domain, there has been a decrease in performance in other domains at varying pruning rates.
This outcome is consistent with expectations because parameter conflicts are a common issue with model merging, invariably resulting in performance degradation.

\section{Conclusions}
In this study, we introduce a pruning method called DP, which is an improved approach based on amplitude pruning to enhance performance at higher pruning rates.
Subsequently, we propose DPA, which focuses on dynamically amplifying partitions of parameters based on their varying levels of importance.
Using DPPA, we address the challenge of model merging in complex fine-tuned models.
The experimental results show that our approach retains only 20\% of the specific domain parameters, yet achieves comparable performance to other methods that retain 90\% of the specific domain parameters.
Furthermore, our method also achieves a significant improvement of nearly 20\% in model merging.
Additionally, we investigate the underlying reasons behind the effectiveness of our proposed method.

\section*{Limitations}
Our method performs less effectively than DARE on fine-tuned models with minimal differences compared to the original model.

DAP requires a longer time to find the optimal ratio.

While it mitigates parameter conflicts in model fusion, there still remains the issue of performance degradation.

\bibliography{custom}

\appendix

\begin{table}
\centering
\small 
\resizebox{0.5\textwidth}{!}{
\begin{tabular}{lccccc}
\toprule
{\textbf{Sparse ratio}} &{Magnitude} &{OWL}  & {DP} & {DARE} \\
\midrule
{gsm8k} \\
{0.1}&{0.59893859}&{0.595905989}&{0.589082638}&{0.587566338} \\
{0.2}&{0.593631539}&{0.592873389}&{0.59893859}&{0.585291888} \\
{0.3}&{0.590598939}&{0.589082638}&{0.594389689}&{0.586808188} \\
{0.4}&{0.578468537}&{0.579984837}&{0.588324488}&{0.567096285} \\
{0.5}&{0.584533738}&{0.589840788}&{0.587566338}&{0.563305534} \\
{0.6}&{0.578468537}&{0.574677786}&{0.570128886}&{0.557240334} \\
{0.7}&{0.546626232}&{0.542835481}&{0.545109932}&{0.558756634} \\
{0.8}&{0.501137225}&{0.495072024}&{0.489006823}&{0.53525398} \\
{0.9}&{0.343442002}&{0.342683851}&{0.351781653}&{0.498104625} \\
\midrule
{MATH} \\
{0.1}&{0.1208}&{0.122}&{0.129}&{0.1236} \\
{0.2}&{0.1218}&{0.1212}&{0.1232}&{0.1298} \\
{0.3}&{0.125}&{0.1232}&{0.1238}&{0.1274} \\
{0.4}&{0.1262}&{0.1258}&{0.1276}&{0.1264} \\
{0.5}&{0.122}&{0.125}&{0.1248}&{0.1216} \\
{0.6}&{0.1254}&{0.124}&{0.1194}&{0.1184} \\
{0.7}&{0.1176}&{0.1148}&{0.1142}&{0.1134} \\
{0.8}&{0.0996}&{0.0934}&{0.095}&{0.111} \\
{0.9}&{0.0646}&{0.0664}&{0.0668}&{0.0842} \\
\midrule
{FiQA\_SA} \\
{0.1}&{0.608510638}&{0.595744681}&{0.595744681}&{0.629787234} \\
{0.2}&{0.612765957}&{0.642553191}&{0.629787234}&{0.621276596} \\
{0.3}&{0.629787234}&{0.646808511}&{0.621276596}&{0.634042553} \\
{0.4}&{0.629787234}&{0.621276596}&{0.629787234}&{0.625531915} \\
{0.5}&{0.582978723}&{0.561702128}&{0.34893617}&{0.561702128} \\
{0.6}&{0.595744681}&{0.540425532}&{0.54893617}&{0.685106383} \\
{0.7}&{0.540425532}&{0.510638298}&{0.195744681}&{0.587234043} \\
{0.8}&{0.519148936}&{0.557446809}&{0.493617021}&{0.570212766} \\
{0.9}&{0.365957447}&{0.395744681}&{0.438297872}&{0.574468085} \\

\bottomrule
\end{tabular}
}
\caption{
All pruning result for three domain.
}
\label{tab:all}
\end{table}

\begin{table*}
\centering
\small 
\resizebox{1\textwidth}{!}{
\begin{tabular}{lccccccccccc}
\toprule
{\textbf{Model}} &{Min} &{10\%}&{20\%}&{30\%}&{40\%}&{50\%}&{60\%}& {70\%}&{80\%}&{90\%} & {Max} \\
\midrule
{Math-Dense} & {-0.0173} & {-0.0011} &{-0.0007}&{-0.0004}&{ -0.0002}&{ 1.175e-08}&{0.0002}&{0.0004}&{ 0.0007} & {0.0011} & {0.0201} \\
{Fin-Dense} & {-0.0261} & {-0.0016} &{-0.0010}&{-0.0006}&{-0.0003}&{0.0}&{0.0003}&{0.0006}&{ 0.0010} & {0.0016} & {0.0201} \\
{Law-Dense} & {-0.0218} & {-0.0015} &{-0.0010}&{-0.0006}&{-0.0003}&{0.0}&{0.0003}&{0.0006}&{0.0010} & {0.0015} & {0.0202} \\
\bottomrule
\end{tabular}
}
\caption{
The offset of different models from the base model at different position proportions.
}
\label{tab:full_amplitude}
\end{table*}
\begin{table}
\centering
\small 
\resizebox{0.5\textwidth}{!}{
\begin{tabular}{lccccc}
\toprule
{\textbf{Sparse ratio}} &{Magnitude} &{OWL}  & {DP} & {DARE} \\

\midrule
{FPB} \\
{0.1}&{0.642268041}&{0.631958763}&{0.58556701}&{0.62371134} \\
{0.2}&{0.620618557}&{0.616494845}&{0.611340206}&{0.634020619} \\
{0.3}&{0.597938144}&{0.608247423}&{0.628865979}&{0.627835052} \\
{0.4}&{0.610309278}&{0.609278351}&{0.601030928}&{0.644329897} \\
{0.5}&{0.590721649}&{0.57628866}&{0.605154639}&{0.611340206} \\
{0.6}&{0.597938144}&{0.579381443}&{0.579381443}&{0.615463918} \\
{0.7}&{0.534020619}&{0.550515464}&{0.537113402}&{0.607216495} \\
{0.8}&{0.460824742}&{0.477319588}&{0.471134021}&{0.586597938} \\
{0.9}&{0.387628866}&{0.38556701}&{0.416494845}&{0.567010309} \\
\midrule
{UNFAIR\_ToS} \\
{0.1}&{0.191860465}&{0.238372093}&{0.26744186}&{0.203488372} \\
{0.2}&{0.284883721}&{0.279069767}&{0.186046512}&{0.191860465} \\
{0.3}&{0.25}&{0.261627907}&{0.209302326}&{0.238372093} \\
{0.4}&{0.244186047}&{0.220930233}&{0.25}&{0.180232558} \\
{0.5}&{0.197674419}&{0.209302326}&{0.197674419}&{0.203488372} \\
{0.6}&{0.279069767}&{0.244186047}&{0.209302326}&{0.226744186} \\
{0.7}&{0.209302326}&{0.23255814}&{0.261627907}&{0.220930233} \\
{0.8}&{0.186046512}&{0.25}&{0.244186047}&{0.13372093} \\
{0.9}&{0.215116279}&{0.26744186}&{0.255813953}&{0.145348837} \\
\midrule
{SCOTUS} \\
{0.1}&{0.216666667}&{0.233333333}&{0.233333333}&{0.3} \\
{0.2}&{0.316666667}&{0.283333333}&{0.283333333}&{0.266666667} \\
{0.3}&{0.283333333}&{0.25}&{0.283333333}&{0.266666667} \\
{0.4}&{0.266666667}&{0.316666667}&{0.35}&{0.25} \\
{0.5}&{0.25}&{0.233333333}&{0.35}&{0.166666667} \\
{0.6}&{0.316666667}&{0.35}&{0.3}&{0.116666667} \\
{0.7}&{0.35}&{0.35}&{0.35}&{0.233333333} \\
{0.8}&{0.316666667}&{0.283333333}&{0.25}&{0.216666667} \\
{0.9}&{0.15}&{0.25}&{0.216666667}&{0.15} \\

\bottomrule
\end{tabular}
}
\caption{
All pruning result for three domain.
}
\label{tab:2}
\end{table}
\section{Main Result of Various Pruning Methods on Specific Tasks}
\label{sec:specific_tasks}
We presented all pruning result in Table~\ref{tab:all} and Table~\ref{tab:2}.

\section{The Offset of Models}
\label{sec:model_offset}
We presented ten different percentage values in Tabel~\ref{tab:full_amplitude}.

\end{document}